# Evaluation of Machine Learning Methods to Predict Coronary Artery Disease Using Metabolomic Data

Henrietta FORSSEN[a*], Riyaz PATEL[b,c*], Natalie FITZPATRICK[b,c], Aroon HINGORANI[d], Adam TIMMIS[e], Harry HEMINGWAY[b,c], Spiros DENAXAS[b,c]

[a]*Department of Computer Science, UCL*
[b]*Institute of Health Informatics, UCL*
[c]*Farr Institute of Health Informatics Research, UCL*
[d]*Institute of Cardiovascular Sciences, UCL*
[e]*NIHR Cardiovascular BRU, Barts*

**Abstract.** Metabolomic data can potentially enable accurate, non-invasive and low-cost prediction of coronary artery disease. Regression-based analytical approaches however might fail to fully account for interactions between metabolites, rely on a priori selected input features and thus might suffer from poorer accuracy. Supervised machine learning methods can potentially be used in order to fully exploit the dimensionality and richness of the data. In this paper, we systematically implement and evaluate a set of supervised learning methods (L1 regression, random forest classifier) and compare them to traditional regression-based approaches for disease prediction using metabolomic data.

**Keywords.** coronary artery disease, random forest, machine learning, EHR

## 1. Introduction

Coronary artery disease (CAD) is one of the leading causes of morbidity and mortality worldwide [1]. Definitive diagnosis is by coronary angiography, an invasive procedure that can lead to severe complications [2] or by additional often costly imaging techniques. Non-invasive blood testing, using circulating metabolites [3] [4], could potentially minimize unnecessary tests and predict CAD with higher accuracy. Previous research however has been mainly restricted to classical regression-based methods [5] [6] and potentially fails to fully exploit the dimensionality and richness of the data by accounting for interactions between metabolites. Supervised machine learning (ML) methods might be better-placed to address these challenges but have yet to be systematically evaluated in this context. Our aims were to a) investigate and evaluate supervised ML methods for CAD prediction using metabolomics data and b) compare their accuracy with traditional regression-based approaches.

## 2. Background

Metabolites are small molecules produced during metabolism or generated by microbes within the body [7]. Metabolites are the end-products of gene expression, a process closely related to protein/enzymatic reactions and therefore potentially offer a direct molecular reflection of the cellular milieu that leads to pathophysiological changes. Circulating metabolites may help predict the presence of CAD by firstly identifying metabolic disturbances, relevant for atherosclerosis (e.g. diabetes and insulin resistance [8] [9]). Additionally, since atherosclerosis occurs at the blood-vessel wall interface, blood metabolite measurements could plausibly directly reflect this chronic process and help predict CAD existence and stability. However, for selected metabolites studied to date, the incremental predictive utility over routine clinical assessments has been modest and restricted to a few candidates measured using non-scalable methods. Recent high-throughput, low-cost and high-dimensional methods [4] (e.g. nuclear magnetic resonance spectroscopy), have re-invigorated hope for using metabolic signatures for CAD prediction but analyses of these complex data present new challenges before realized.

ML techniques are data-driven approaches designed to discover statistical patterns in large high-dimensional multivariate data and have been previously used for creating accurate risk prediction models [10]. Supervised ML methods are a set of techniques which aim to infer a function from a labelled training dataset which can predict the class of future input vectors. We evaluated penalized logistic regression and random forest to assess the predictive performance of metabolites on CAD in a contemporary cohort of patients referred to hospital for chest pain investigation or planned coronary angiography.

## 3. Methods

### 3.1. Case and exposure definitions

We used data from the Clinical Cohorts in Coronary disease Collaboration (4C) study (n=3409) which recruited patients with acute or stable chest pain from four UK NHS hospitals [11]. Patients consented to having their EHR extracted and provided blood samples. We defined presence of CAD as a >50% stenosis [12] occurring in ≥1 coronary arteries using data from: a) coronary angiography reports and b) EHR evidence of previous coronary revascularization procedures (Percutaneous Coronary Intervention, Coronary Artery Bypass Graft) recorded in EHR. Participants in whom CAD could not be ascertained were excluded. For each participant, 256 metabolites were quantified using an NMR technique. Full details have been published elsewhere [4] [11]. Missing metabolite values were imputed and zero mean unit standardized by multiple imputation [13] (predictive mean matching [14]) and standardized to zero mean unit variance by first subtracting the means and dividing by the standard deviations. Data were randomly split into training and test subsets using a 3:1 ratio.

### 3.3 Statistical methods

We performed logistic regression on each of log+1-transformed metabolite values adjusting for known risk factors. We derived principle component factors for the standardized metabolite values and selected the first six for analyses as they accounted for >95% of the data variability. We then performed logistic regression on each of the Principal Component Analysis (PCA)-derived metabolite factors, and multiple logistic regression including all six. Adjusted (age, sex, use of statins, hypertension) and unadjusted models were Bonferroni corrected (p<0.05). We performed penalized logistic regression using the Lasso penalty which was defined as the lowest error obtained from a 50-fold cross-validation. We trained a random forest classifier using Gini impurity and 5,000 trees per ensemble. Initial cross-validation was conducted on the training set for both the proportion of variables used per tree as well as the maximum tree depth. A second cross-validation was conducted on the number of variables alone, whilst allowing trees to grow to their maximum depth. This removed the uncertainty of tuning a second parameter, and the possible increase in variance due to increased depth was considered well counterbalanced by using a very large number of ensemble trees. Final predictions were the average individual pooled predictions [13] [14] across imputed datasets and evaluated by calculating the percentage of correct predictions, ROC curves and AUC.

## 4 Results

We identified 1474 patients with metabolomics in whom CAD was ascertained (Table 1).

Table 1. Summary of study population

| Clinical Characteristics | | Clinical Characteristics | |
|---|---|---|---|
| Men (%) | 1106 (78%) | Statin use (%) | 447 (30%) |
| Age (Years) | 62.4± 11.6 | Diabetes (%) | 523 (35%) |
| BMI (kg/m2) | 29.4± 5.3 | Current smoker (%) | 278 (28%) |
| Diagnosed hypertension (%) | 1146 (77%) | CAD present (%) | 1037 (70%) |

### 4.1 Comparison of model predictiveness

In the unadjusted models, the random forest classifier had the highest AUC and accuracy values and highest ROC curve (Figure 1) and both ML models outperformed PCA regression (Table 2). AUC, raw accuracy and PPV were mostly similar across models. All models had significantly higher sensitivities than specificities, but PCA regression had the most extreme values as it predicted the vast majority of positive CAD cases correctly but nearly none of the negative CAD cases. The large disparity in sensitivity and specificity for the two other two models shows that they failed to accurately distinguish between disease states. When adjusting for confounders, PCA regression had the best accuracy but higher AUC and accuracy values compared to unadjusted models were observed in all models.

Table 2. Adjusted/unadjusted prediction results; highest AUC values highlighted.

| Model | Accuracy | AUC | Sensitivity | Specificity | PPV | NPV |
|---|---|---|---|---|---|---|
| PCA regression | 0.686 | 0.625 | 0.984 | 0.026 | 0.691 | 0.429 |
| PCA regression adjusted | 0.759 | 0.767 | 0.957 | 0.322 | 0.757 | 0.771 |
| L1 regression | 0.688 | 0.663 | 0.882 | 0.261 | 0.725 | 0.550 |
| L1 regression adjusted | 0.767 | 0.765 | 0.949 | 0.339 | 0.760 | 0.750 |
| Random forest | 0.713 | 0.675 | 0.941 | 0.209 | 0.724 | 0.615 |
| Random forest adjusted | 0.732 | 0.711 | 0.937 | 0.278 | 0.741 | 0.667 |

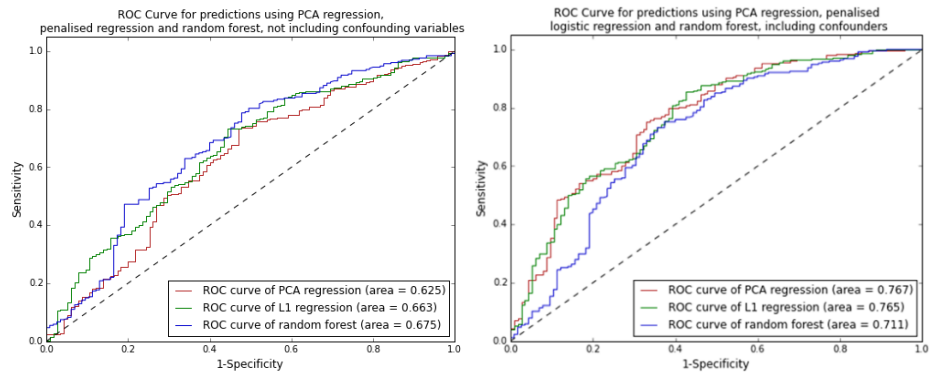

**Figure 1.** Unadjusted (left) and adjusted (right) ROC curve

*4.2 Model compositions and predictive metabolites*

**Logistic regression:** In individual metabolites, strong CAD associations after Bonferroni correction were primarily related to lipids, (e.g. HDL and VLDL), apolipoprotein A-I, the ratio of triglycerides to phosphoglycerides and the ratio of omega-6, monosaturated, polysaturated values to total fatty acids. With logistic regression on individual PCA-derived factors, the first PCA factor (VLDL, ratio of apolipoprotein B to apolipoprotein A-I, ratio of apolipoprotein B to apolipoprotein A-I, 0.404 variance) remained statistically significant. When using all factors and adjusting for confounders, the first and second factors (IDL and LDL, 0.165 variance) were statistically significant.

**Penalized logistic regression:** In the unadjusted models, the ratio of apolipoprotein B to apolipoprotein A-I was found to have the largest, statistically significant negative association with presence of CAD. This was followed by cholesterol esters in small LDL which had a large positive association with presence of CAD. Saturated fatty acids also had a large negative association, whilst phospholipids in chylomicrons and extremely large VLDL had a large positive association. Overall ~70 predictive metabolites where included in each model with 117 metabolites included in at least one of the models used to average the prediction. This suggests that whilst excluding confounders, it is difficult to select a small profile of metabolites to accurately predict the presence of absence of CAD using penalized regression. When adjusting for confounders, substantially fewer metabolites were selected; the ratio of monounsaturated fatty acids to total fatty acids and triglycerides to total lipids ratio in IDL had the largest statistically significant positive association while glutamine and acetoacetate had negative associations.

**Random forest:** In the unadjusted classifier, creatinine was the most strongly significant metabolite followed by triglycerides to total lipids ratio in IDL, phenylalanine, albumin and lactate. Similar predictors were observed in the adjusted models with age being the most significant component followed by creatinine, triglycerides to total lipids ratio in IDL, phenylalanine, albumin and lactate. Similar metabolite profiles for adjusted/unadjusted models suggest that random forest does not incorporate the additional information of confounding variables as well as the other models.

## 5  Concluding discussion

While ML approaches predicted presence/absence of CAD in the unadjusted models (using metabolite data only) with high accuracy/sensitivity, when adjusting for confounders they were outperformed by PCA regression in terms of ROC AUC and accuracy suggesting that a small number of metabolites can potentially be included in prediction models. Multiple individual metabolites that were found statistically significant are in agreement with previous literature and our pathological understanding of CAD and its development. Among these, the atherogenic lipid particles such as LDL are known to be causally related to atherosclerosis, while others such as creatinine reflect renal function and are also established markers of CAD risk. Several other metabolites have no previous robust association with CAD including phenylalanine and lactate and represent potentially novel avenues for investigation. However in seeking a metabolic signature to predict CAD, ML models suffered from low specificity.

This exploratory analysis has identified and exemplified the value of ML models for CAD prediction using high-dimensional data, and shown that accuracy of traditional regression-based approaches can be surpassed. Nonetheless further research is required before these methods can be translated into clinical solutions.